\documentclass[12pt, authoryear]{article}

\usepackage{graphicx}
\usepackage{epsfig}
\usepackage{amssymb}
\usepackage{amsmath, amsthm}
\usepackage{soul}
\usepackage{xcolor}
\usepackage{accents}

\usepackage{algorithm}
\usepackage{algorithmic}
\usepackage{enumitem}
\usepackage{multicol}
\usepackage{indentfirst}
\usepackage{breqn}
\usepackage{natbib}

\newcommand{\pr}{{\mathbb{P}\,}}

\newcommand{\ex}{{\mathbb{E}\,}}

\newcommand{\as}{\mathcal{A}}

\newcommand{\probP}{\text{I\kern-0.15em P}}

\graphicspath{ {./res_images/} }

\title{Reinforcement Learning in \\ Credit Scoring and Underwriting}

\author{Seksan Kiatsupaibul
\thanks{Department of Statistics, Chulalongkorn University, Bangkok 10330, Thailand}
\and
Pakawan Chansiripas\footnotemark[1]
\and
Pojtanut Manopanjasiri\footnotemark[1]
\and 
Kantapong Visantavarakul\footnotemark[1]
\and
Zheng Wen
\thanks{Deepmind, Mountain View, CA 94043, USA}
}
 
\begin{document}

\maketitle

\begin{abstract}
This paper proposes a novel reinforcement learning (RL) framework for credit underwriting that tackles ungeneralizable contextual challenges. We adapt RL principles for credit scoring, incorporating action space renewal and multi-choice actions. Our work demonstrates that the traditional underwriting approach aligns with the RL greedy strategy. We introduce two new RL-based credit underwriting algorithms to enable more informed decision-making. Simulations show these new approaches outperform the traditional method in scenarios where the data aligns with the model. However, complex situations highlight model limitations, emphasizing the importance of powerful machine learning models for optimal performance. Future research directions include exploring more sophisticated models alongside efficient exploration mechanisms.
\end{abstract}


\section{Introduction}
Credit scoring has been a staple of machine learning applications for decades~\citep{Phillips2018, dastile2019}. 
As businesses have adopted data-driven platforms and accumulated large datasets, this application has become increasingly prevalent and crucial not only for financial institutions, but also for other institutions that extend credit. However, treating credit scoring as an independent step within the credit underwriting process can be inefficient. While the model is being developed to fit the business needs, an institution must underwrite loans with a less accurate model or even without a model in order to obtain a labeled dataset for model construction. This can be costly for the institution, as they may accept a large number of potentially risky loans, leading to higher default rates and losses. Therefore, a holistic view that considers the entire credit scoring and underwriting process, from data collection to model development to loan approval, is necessary to maximize total profit, which includes the cost incurred by underwriting mistakes. A bandit framework in reinforcement learning offers a promising solution for achieving this goal.

A bandit framework in reinforcement learning is a promising approach to achieve this goal. A bandit framework allows the institution to learn and adapt to the characteristics of borrowers while maximizing long-term accumulated profit, taking into account the risk of default. In a bandit framework, an agent is faced with a series of decisions, each of which has a potential reward. The goal is to make a series of decisions that maximize the total cumulative reward \citep{Sutton2018, Russo2018}. When applied to credit scoring and underwriting, an agent (or lender) sequentially decides which borrowers to underwrite. The reward of each decision is either the interest earned from the loan if the borrower repays, or the loss of capital if the borrower defaults. Since the objective is to maximize the total reward, the model captures all costs from the entire credit scoring and underwriting process.

Traditionally, credit scoring has been treated as a standalone operation under the underwriting process. When viewed separately, 
credit scoring can be framed as a statistical prediction or machine learning problem. 
Logistic regression is among the most widely used machine learning models for credit scoring due to its simplicity, robustness, and interpretability. In this study, we propose a unified approach to credit scoring and credit underwriting by modeling the 
whole process as a bandit problem within the reinforcement learning framework. 
Accordingly, the logistic bandit model serves as the primary model for our investigation~\citep{faury2020abeille, Slivkins2019}. 
We focus on the ungeneralizable contextual logistic bandit model, as borrowers often belong to diverse groups with distinct characteristics~\citep{chapelle2011, langford2007}. In this scenario, a more sophisticated reinforcement learning agent is necessary to effectively navigate this 
challenging environment.

This paper is organized as follows.  In Section~2,  we model the combination of credit scoring and credit underwriting as a 
reinforcement learning environment. We consider different scenarios, from a simple environment to more practical ones.  In Section~3, we design reinforcement learning agents to handle the introduced credit scoring and underwriting environments.  In Section~4, we provide the numerical results showing the performances of different reinforcement learning agents when applied to different credit scoring and underwriting environments. We also discuss the results. In Section~5, we make a conclusion.

\section{Credit Scoring and Underwriting Environment}\label{sec:env}

In lending operations, credit scoring is a traditional step within the credit underwriting process. However, it's often operated as a standalone component. While credit scoring primarily evaluates the risk associated with each loan applicant, credit underwriting involves making decisions on granting loans based on these scores. The main goal of credit underwriting is to generate revenue from interest payments. If the credit scoring model is constructed independently using a fixed, un-updated dataset, combining the two processes doesn't yield any benefits. In such a setting, the optimal underwriting rule is simply to grant loans to those with the best credit scores.

In practice, the default outcomes of loans approved through underwriting are fed back into the credit scoring model construction as an updated dataset. This creates interactions between the credit scoring and credit underwriting processes. It is important to note that underwriting loans solely based on the best scores may no longer be optimal. Underwriting loans to those with lower scores may yield better cumulative revenue in the long run if the information obtained from the defaults of these borrowers helps improve the existing credit scoring model for better future predictions.

\begin{figure}[h!]
\begin{center}
\includegraphics[scale=0.7]{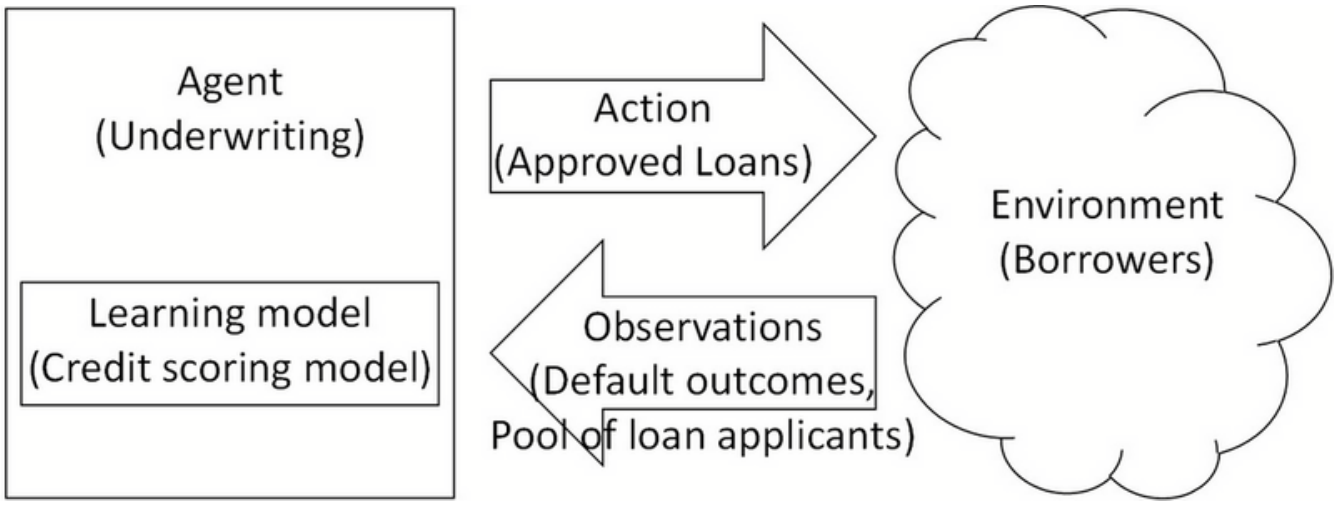}
\caption{Reinforcement learning components and their credit underwriting counterparts.}\label{fig:rlcredit}
\end{center}
\end{figure}

This section proposes a unified approach that treats credit scoring and credit underwriting as a single process using a 
reinforcement learning model. 
A reinforcement learning model involves an agent interacting with an environment in repeated cycles. The cycle begins with the agent making decisions to the environment.  The environment responds with observations and determines rewards for the agent. Based on the history of decisions and observations, the agent then makes new decisions to achieve the highest cumulative rewards in the long run.  Figure ~\ref{fig:rlcredit} illustrates typical reinforcement components and their corresponding credit underwriting counterparts. 

\ \\

In the proposed reinforcement learning model for credit scoring and underwriting, 
the agent represents the lender along with its decision rules. 
The lender processes the history of lending decisions and observations of default outcomes, resulting in a credit scoring model. Based on this model, the lender makes decisions to underwrite loans. The environment consists of the pool of loan applicants or potential borrowers and their creditworthiness. After the lender decides who to underwrite, the approved borrowers generate default-non-default observations based on their creditworthiness. The reward to the lender is the net profit determined by the capital lost in case of defaults and the interest earned in case of non-defaults. The lender then updates the credit scoring model using the new default-non-default observations and adds the new observations into the history of past decisions and observations.  
The process repeats.  The objective is to design a lending agent (decision rule) that can achieve the highest cumulative reward 
or total profit in the long run.

In each cycle of credit scoring and underwriting, the environment is assumed to be a pool of $K$ loan applicants.  
Each applicant is identified by their characteristics summarized by a vector of features, $x\in\Re^d$. 
If granted loans, the loan applicants default independently from one to another with probability of default 
assumed to be a function of $x$.  
These loan applicants are drawn uniformly from $M$ groups of potential borrowers 
where the applicants from different groups have different default patterns.
Conditional on the same $x$, two applicants from different groups with the same $x$ have different probabilities of default.  
In each cycle of credit scoring and underwriting, the lender evaluates the probabilities of default or the credit scores of 
$K$ loan applicants and grants the loans to a subset of size $N<K$ of the applicants.

We examine two scenarios of the credit scoring and underwriting environment, a simple one and a practical one. The simple scenario involves the direct application of the simple reinforcement learning environment where the pool of loan applicants is fixed, and the lender grants a loan to one applicant in each round of underwriting. However, this scenario lacks two crucial practical characteristics. In practice, the group of loan applicants is constantly changing, and the lender grants multiple loans in each round of underwriting. The practical scenario  incorporates these two practical characteristics into consideration. 
The two lending scenarios are the following.

\begin{enumerate}
\item {\it Single underwriting with a fixed pool of loan applicants (SF):} The pool of $K$ loan applicants is fixed 
for every round of credit scoring and underwriting and only one  applicant ($N=1$) is granted the loan in each round.
\item {\it Multiple underwriting with a renewed pool of loan applicants (MR):} The pool of $K$ loan applicants is renewed 
for each round of credit scoring and underwriting and more than one applicants ($1<N<K$) are granted the loans in each round.
\end{enumerate}

In the next section, we will discuss the lending decision rules for each of these two scenarios.

\section{Agents Design}

Traditional credit scoring models, such as logistic regression, while informative, lack the ability to optimize for specific business goals and adapt to new information. Here, we explore the potential of reinforcement learning agents to enhance credit scoring. These agents incorporate decision-making capabilities that balance the short-term learning cost with the long-term cumulative reward. We consider three types of reinforcement learning agents: greedy agent (GRE), Thompson sampling agent (THS), and information-directed sampling agent (IDS). Notably, all three agents leverage a logistic regression model as their learning models, 
but they differ in how they use this model to score applicants and to make lending decisions.

\begin{algorithm}
\caption{Underwriting Algorithm}
\label{algo:underwriting}
\begin{algorithmic}[1]
\STATE \textbf{input} Prior distribution $\mathcal{N}(\mu,\Sigma)$, 
\STATE \textbf{initialization} $H_0=(\ )$
\FOR{ $t=0,1,\ldots$}
\STATE \textbf{recruiting} Obtain a pool of loan applicants $\as$. 
\STATE \textbf{scoring} Derive credit score $s_a$ for each $a\in\as$
\STATE \textbf{making decision} Rank $a\in\as$ ascendingly by the score $s_a$, breaking ties randomly, and 
let $A_t$ be the first $N$ members (top $N$ scorers).  Extend loans to each $a\in A_t$.
\STATE \textbf{receiving feedback} Observe $O_{t+1}$, the default outcomes of approved loans in $A_t$.
\STATE \textbf{updating} $H_{t+1}\leftarrow \text{append}(H_t, A_t, O_{t+1})$
\ENDFOR 
\end{algorithmic}
\end{algorithm}

Three reinforcement learning agents, each with a unique scoring method, 
utilize the same core underwriting algorithm outlined in Algorithm~\ref{algo:underwriting}. 
This algorithm functions iteratively: at each time step, the agent recruits a pool of loan applicants $\as$ where $|\as|=K$.  
Each applicant $a\in\as$ is characterized by two data vectors: 
a context vector $\phi_a \in\Re^M$ and a feature vector $x_a\in\Re^d$. 
The context vector,
\[\phi_a=[\phi_{a,1},\ldots, \phi_{a,M}]^\top,\]  
is a one-hot encoded group indicator, meaning all elements are zero except for 
a single element, with value one, representing the applicant's group affiliation. 
The feature vector, 
\[x_a = [x_{a,1},\ldots, x_{a,d}]^\top,\]
on the other hand, captures the applicant's individual characteristics relevant to creditworthiness. 
Once the applicant pool is formed, the agent employs its specific scoring method 
to evaluate and rank each applicant.  Lower scores indicate a lower risk of loan default.
Subsequently, the agent grants loans to the top scorers (applicants with lowest scores), 
observes their repayment outcomes (defaults or non-defaults), 
and updates its history that is the record of actions and outcomes.

The logistic bandit agent operates under the assumption that each borrower's default probability follows a logistic regression model.  
Here, we make the additional assumption that default patterns are specific to $M$ borrower groups,  
and knowledge gained from one group cannot be applied to others.  
Consequently, the logistic bandit agent assumes that each borrower group has a unique set of logistic regression model parameters.  This specific scenario falls under the category of ungeneralizable contextual logistic bandits. 
Let $Y_a$ denote an indicator variable that takes value 1 if borrower $a$ defaults on the loan and 0 if they pay back the loan. 
An ungeneralizable contextual logistic bandit agent assumes the probability of default follows this logistic regression equation:
\begin{equation}\label{eqn:logit}
\pr(Y_a=1) = \frac{\exp\left(\phi_a^\top\Theta x_a\right)}{1+\exp\left(\phi_a^\top\Theta x_a\right)},
\end{equation}
In this equation, the model parameters $\Theta$ is an $M\times d$ matrix containing the logistic regression model parameters. 
From (\ref{eqn:logit}), each row of $\Theta$ corresponds to the parameter vector for a specific borrower group.

Extending the loan to applicant $a$, the agent receives a return of $v$ if $a$ pays back the loan and incurs a 
loss of $l$ if $a$ defaults on the loan.
Therefore, given $\Theta$, the expected reward for the agent if they extend the loan to $a$ is 
\begin{eqnarray}\label{eqn:reward}
\ex\left[\text{Reward}_a\mid\Theta\right] & = & v\left(1-\pr(Y_a=1)\right) - l\pr(Y_a=1) \\
 & = & v -  (v+l)\pr(Y_a=1).
\end{eqnarray}
As a result, given $\Theta$, ranking applicants descendingly by the reward is equivalent to 
ranking them ascendingly by the probability of default, and 
is also equivalent to ranking them by the log odd $\phi_a^\top\Theta x_a$.

All three agents employ Bayesian updating to learn about the model parameters. 
Each agent maintains a history, denoted by $H_t$,
\[H_t = (A_0, O_1, A_1, O_2, \ldots, A_{t-1}, O_t),\] 
which tracks the sequence of its actions and the corresponding observations.  
Note that $A_t$ is a collection of top scorers that contains $N$ loan applicants and 
$O_t$ is the collection of default-non-default outcomes of multiple applicants, $Y_a, a\in A_t$.
This history is crucial for estimating the probability of the model parameters $\Theta$ given the observed data.
Equation (\ref{eqn:posterior}) shows this probability, formally called the posterior distribution. 
\begin{equation}\label{eqn:posterior}
f(\Theta\mid H_t) \propto \pr(H_t\mid\Theta)f(\Theta) = 
\left[\prod_{\tau=0}^{t-1} \pr\left(O_{\tau+1}\mid A_\tau, \Theta\right)\right] f(\Theta),
\end{equation}
where $f(\Theta)$ is the prior distribution and $\pr\left(O_{\tau+1}\mid A_\tau, \Theta\right)$ is given in (\ref{eqn:logit}). 
By analyzing the posterior distribution, the agents can gain insights into the true model parameters. 
However, the key difference among the agents lies in how they leverage this posterior information 
within their unique scoring methods.

{\flushleft\emph{Greedy Agent (GRE)}:}\\

A simple approach involves a greedy agent. 
This agent employs a typical credit scoring approach with a scoring model that is sequentially updated based on observations from actions taken using previous models.  It is purely exploitative, aiming to maximize reward from existing information 
without exploring potentially valuable alternative choices to gather additional information.

At time step $t$, let $\hat{\Theta}_t$ be a point estimate, which is the maximum a posteriori (MAP) estimate according to 
a Bayesian framework.  The greedy agent chooses the action that maximizes the expected reward with respect to $\hat{\Theta}_t$.  
That is the greedy agent extends the loans to the top $N$ scorers, whose score is defined as the probability of non-default.  
The {\bf scoring} step in Algorithm~\ref{algo:underwriting} for the greedy agent becomes the following.

{\flushleft 5:\hspace{0.5cm} {\bf scoring} Let $\hat{\Theta}_t$ be MAP estimate of $\Theta$ given $H_t$.} 
Let the score of applicant $a$ be
\begin{equation}\label{eqn:greedyscore}
s_a = \frac{\exp\left(\phi_a^\top\hat{\Theta}_t x_a\right)}{1+\exp\left(\phi_a^\top\hat{\Theta}_t x_a\right)}.
\end{equation}

{\flushleft\emph{Thompson Sampling Agent (THS)}:}\\

The greedy agent suffers from an exploration-exploitation dilemma. 
However, it may exploit a seemingly profitable strategy based on existing information, neglecting potentially better options. Thompson sampling offers an alternative by maintaining probability distributions over potential loan outcomes for different applicants. It balances exploiting known good options with exploring less certain ones, ultimately aiming for a diversified and continuously improving decision-making process~\cite{Russo2018}.

At time step $t$, a Thompson sampling agent samples a particular action according to the probability that the action 
offers the highest rewards.  The probability of choosing an action is driven by the posterior probability distribution of 
the model parameters. Therefore, the agent can also sample the parameters according to the posterior distribution and 
choose the action that offers the highest expected rewards or, equivalently, 
the highest probability of non-default with respect to those sampled parameters.
The {\bf scoring} step in Algorithm~\ref{algo:underwriting} for the Thompson sampling agent becomes the following.

{\flushleft 5:\hspace{0.5cm} {\bf scoring} Sample $\tilde{\Theta}_t$ from $f(\Theta\mid H_t)$.} 
Let the score of applicant $a$ be
\begin{equation}\label{eqn:thompsonscore}
s_a = \frac{\exp\left(\phi_a^\top\tilde{\Theta}_t x_a\right)}{1+\exp\left(\phi_a^\top\tilde{\Theta}_t x_a\right)}.
\end{equation}

The technical difficulty of implementing (\ref{eqn:thompsonscore}) lies in sampling from the posterior distribution $f(\Theta\mid H_t)$.  
In the numerical section that follows, we circumvent the problem by using the Laplace approximation that approximates 
the posterior distribution by a normal distribution whose mean vector is MAP estimate of the mean and the covariance matrix is 
the inverse matrix of the observed Fisher information.

{\flushleft\emph{Information-directed Sampling Agent (IDS)}:}\\

While Thompson Sampling encourages exploration, it can become inefficient when valuable domain knowledge is available. Information-directed sampling (IDS) addresses this by incorporating such knowledge into the exploration and decision-making process~\citep{DanielRusso2017}. This targeted exploration allows the agent to learn more efficiently from the underlying structure of the problem.

The key challenge in our credit underwriting lies in the context-specific nature of the data. These contexts are ungeneralizable, meaning the knowledge gained from one situation may not apply to others. Early in the learning process, 
granting loans to applicants from specific contexts can be more informative than focusing on others, 
especially when the agent has limited knowledge about those particular contexts.

Let $I(X; Y)$ denote the mutual information between two random variables $X$ and $Y$.  
We define the information gain at time $t$, denoted by $\Delta_t:\as\to\Re_+$ as the mutual information, 
as the mutual information between the model parameters pertaining to the applicant's group and the default 
outcome of the applicant, given that the loan is extended to the applicant.
\begin{equation}\label{eqn:infogain}
\Delta_t(a) = I(\phi_a^\top\Theta_t; Y_a\mid a\in A_t)
\end{equation}
Define the expected regret $\text{regret}_{t,a}$ as the difference between the expected reward of the top scorer 
and that of applicant $a$ with respect to the posterior distribution at time $t$.

An IDS agent chooses an action that offers the best balance between regret and information gain. 
It prioritizes actions with low regret and a large amount of information.  
The IDS score of an applicant is defined as the ratio between a polynomial of order p in regret and information gain.
The scoring step in Algorithm~\ref{algo:underwriting} for the IDS agent becomes the following.

{\flushleft 5:\hspace{0.5cm} {\bf scoring} Given the posterior distribution of the model parameters, $f(\Theta\mid H_t)$,} 
let the score of applicant $a$ be
\begin{equation}\label{eqn:idsscore}
s_a = \frac{\text{regret}_{t,a}^p}{\Delta_t(a)}.
\end{equation}

The order of the polynomial in Equation (\ref{eqn:idsscore}), denoted by $p$, is left as a tuning parameter. 
In the numerical section's implementation, the posterior distribution $f(\Theta\mid H_t)$ is approximated 
by a normal distribution using the Laplace approximation. The expected regret at time step $t$ can then 
be evaluated by simulation from the approximate distribution of $f(\Theta\mid H_t)$.  
Likewise, the information gain can be derived using the approximate distribution of $f(\Theta\mid H_t)$.

\section{Numerical Results}

This section compares the performance of three reinforcement learning agents for credit underwriting: greedy (GRE), Thompson sampling (THS), and information-directed sampling (IDS). We evaluate them across environments, using the greedy agent as the baseline.

Rationale for the Greedy Baseline:
\begin{itemize}
\item The greedy agent reflects current credit scoring practices, making it a relevant benchmark.
\item Outperformance by other agents suggests potential for improvement in real-world underwriting.
\item The greedy agent demonstrates strong performance in a single context environment, providing a challenging standard for comparison.
\end{itemize}

We evaluate these agents using a simulation framework that captures the key characteristics of different lending environments. 
The simulated environments encompasses two lending scenarios derived from two data generation models. 
These scenarios include single underwriting with fixed applicant pools (SF) and multiple underwriting with renewed pools (MR) - detailed definitions are provided in Section~\ref{sec:env}. 
We consider two data generation models:
\begin{itemize}
\item Logistic regression,
\item Neural network.
\end{itemize}
The first model allows the environments to generate data that aligns with the underlying models used by the reinforcement 
learning agents. 
These settings enable us to observe agent behavior in an idealized case where 
the environment can be approximated well by a linear model.
The second model introduces a mismatch between the environment's data generation process and the internal models of the agents. 
This setting reflects a more challenging scenario where the real-world environment might be much more complex than 
the models traditionally employed in practice.

For all environments, we set the following parameters:
\begin{itemize}
\item Number of contexts, $M=4$,
\item Number of features, $d=10$ (including the intercept in case of the logistic regression environment),
\item Number of loan applicants in the pool per time step, $K=100$,
\item Number of applicants awarded per action: $N=1$ for single underwriting and $N=10$ for multiple underwriting,
\item The return $v=0.2$ and the loss $l=-0.8$.
\end{itemize}

We employ the cumulative regret curve over time steps as the performance measure for our reinforcement learning agents.
Lower cumulative regret signifies better agent performance, with the ideal scenario being a curve that tapers off over time steps. 
This tapering off reflects the agent's learning process, where it converges towards the optimal strategy for each lending environment.

\begin{figure}[h!]
\begin{center}
\includegraphics[scale=0.45]{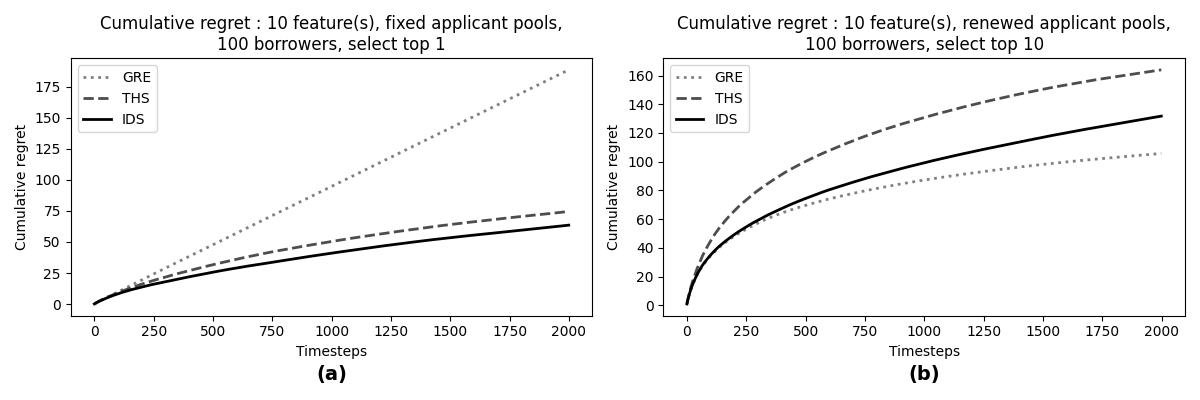}
\caption{The agents' performances under 2 scenarios within the single context environments.  Panel (a) and (b) exhibit 
the cumulative regret curves under the SF and MR scenarios, respectively.}\label{fig:singlecontext}
\end{center}
\end{figure}

Before proceeding with the main experiments, we analyze agent performance in a single context lending environment (e.g., $M=1$). Panel (a) of Figure~\ref{fig:singlecontext} depicts the cumulative regret (lower is better) of the three agents when 
the applicant pool remains fixed, with each agent awarding one loan per time step. As anticipated in a standard logistic bandit setting, 
IDS and THS outperform GRE due to their efficient exploration capabilities.

Panel (b) of Figure~\ref{fig:singlecontext} presents the scenario where the applicant pool is entirely renewed in each time step and 
each of the three agents grants ten loans per time step. Here, the GRE agent 
unexpectedly exhibits the best performance. This is attributable to the constant influx of new applicants, 
which inherently introduces exploration. In this specific setting with a single context, a renewed applicant pool 
and multiple underwriting, 
the GRE agent leverages this inherent exploration to outperform the other agents that are equipped with 
their dedicated exploration mechanisms. 
This finding underscores the importance of the GRE agent as the baseline. 
In the main experiments with multiple contexts, we cannot simply assume that more sophisticated agents 
will automatically surpass the greedy baseline.

In the next subsections, we begin our investigation on the logistic regression environment.  
Subsequently, we will explore the neural network environment.

\subsection{Environment derived from a logistic regression model}

This environment utilizes a logistic regression model to generate borrower default data based on ten features ($d=10$)
representing their characteristics. This alignment between the data generation process and the agents' statistical learning model creates an idealistic scenario for evaluating their performance. We investigate the performance of the three agents in this controlled environment.

\begin{figure}[h!]
\begin{center}
\includegraphics[scale=0.45]{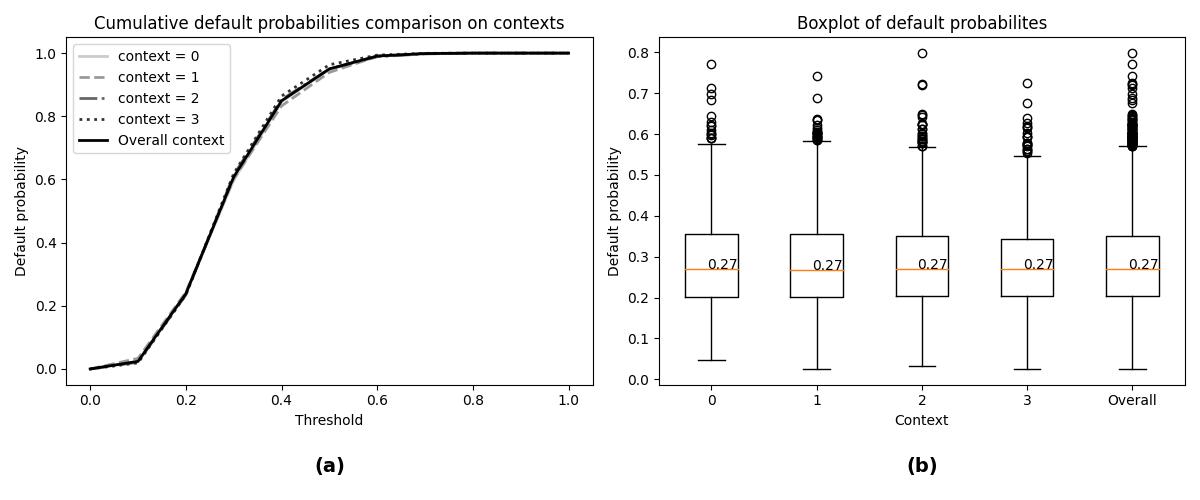}
\caption{The distributions of the default rates by contexts that are simulated under the environment derived from 
the logistic regression data generation model.  Panel (a) illustrates the distributions by the cumulative distribution function.  
Panel (b) summarizes the distributions in terms of box plots.}\label{fig:dfratelogistic}
\end{center}
\end{figure}

The borrower feature vectors are drawn from a ten-dimensional multivariate normal distribution with zero mean and an identity covariance matrix. The logistic regression coefficients, excluding the intercept, are also sampled from the same distribution. The intercept is set to $-1$.  This intercept value controls the mean default rates around $27\%$ for each context.  Each scenario is simulated fifty times to account for variability in the data generation process. The resulting distribution of default rates across these simulations is visualized as a box plot in Figure~\ref{fig:dfratelogistic}.

Based on the mean default rates in 
Figure~\ref{fig:dfratelogistic}, the chosen reward function parameters (return $v=0.2$ and loss $l=-0.8$) 
indicate a negative average net profit for a random loan approval strategy. 
To achieve profitability, the agent must learn a more effective underwriting algorithm that generates a positive net profit.

Scenarios 1 and 2 correspond to the SF and MR environments described in Section~\ref{sec:env}, respectively. The reported performance metrics represent the average outcome across the fifty simulations


{\flushleft \it Scenario 1: Single underwriting with a fixed pool of loan applicants (SF)}\\

We evaluated the performance of the three agents (Greedy (GRE), Thompson Sampling (THS), and Information-Directed Sampling (IDS)) in the standard logistic bandit environment with four ungeneralizable contexts under the SF scenario (refer to Section~\ref{sec:env} for details). In this scenario, the applicant pool remains constant throughout multiple underwriting rounds. In each round, only one applicant receives loan approval.

\begin{figure}[h!]
\begin{center}
\includegraphics[scale=0.35]{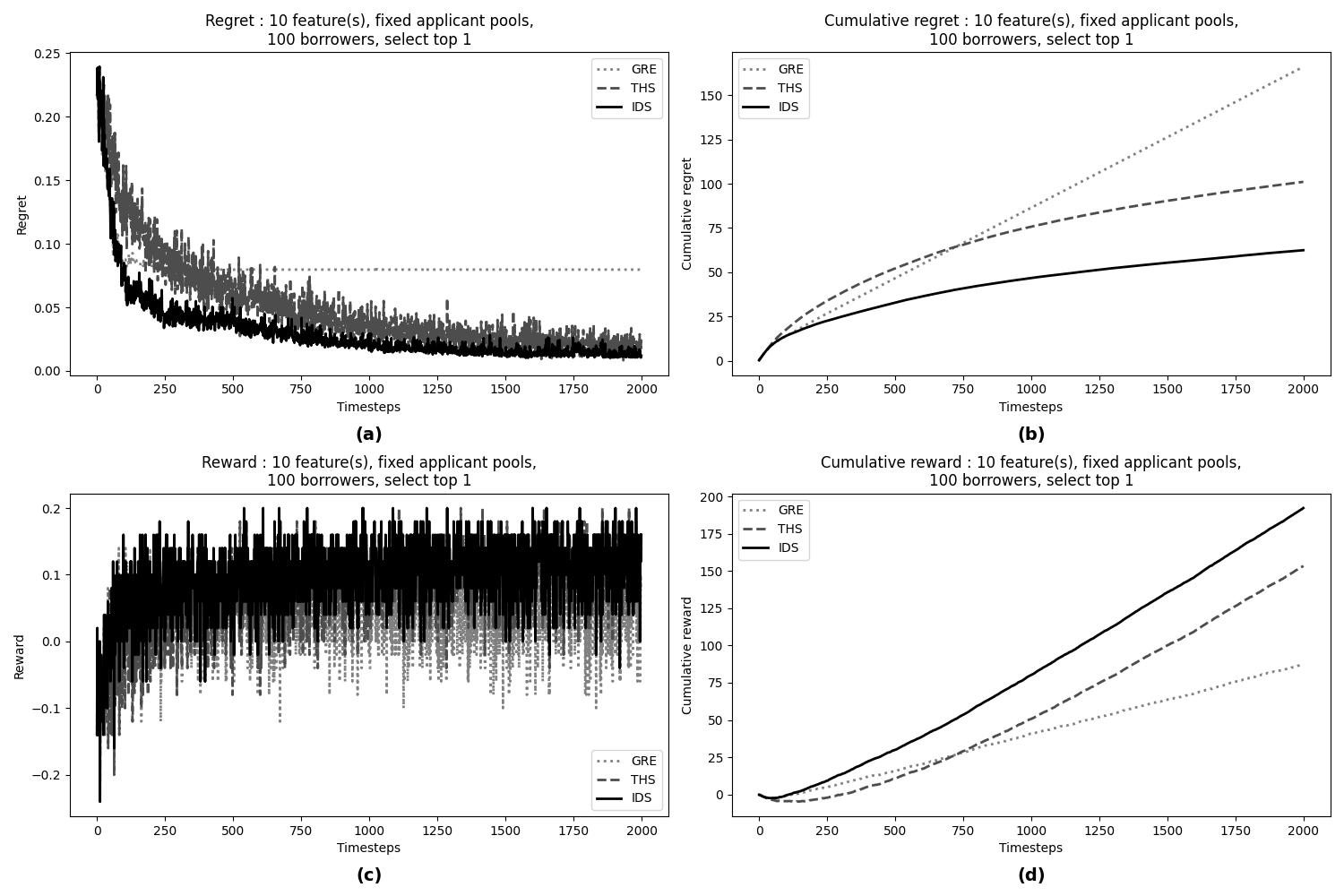}
\caption{The agents' performances with respect to SF scenario (single underwriting, fixed applicant pool) 
within the logistic regression environment.  
Panel (a) and (b) exhibit the regret and the cumulative regret curves.  Panel (c) and (d) exhibit the 
reward and cumulative reward curves.}\label{fig:sf_lr10}
\end{center}
\end{figure}

The results are shown in Figure~\ref{fig:sf_lr10}. The tuning parameter p for the IDS agent was set to 3 for this specific environment and used consistently throughout the experiments. (Lower values on the y-axis of the regret curves indicate better performances).

As anticipated for reinforcement learning agents in a standard logistic bandit environment, those with effective exploration strategies, THS and IDS, outperform the GRE agent. In Figure~\ref{fig:sf_lr10} (a), the regret curves for THS and IDS decrease towards zero, while the GRE agent's curve remains flat at a positive value. This indicates that THS and IDS agents learn the optimal model over time, whereas the GRE agent fails to do so. 

These trends are further confirmed in Figure~\ref{fig:sf_lr10} (b), which displays the cumulative regret curves for all three agents. The curves for THS and IDS plateau over time, indicating their learning effectiveness. Conversely, the GRE agent's curve continues to rise, signifying its inability to learn effectively. Notably, the cumulative regret curves exhibit less noise compared to the standard regret curves, allowing for clearer distinction between the performance of the three agents.

By comparing the cumulative regret curves between the agents with efficient exploration, 
we observe that the IDS agent achieves a lower value than the THS agent. 
This suggests that IDS strikes a more efficient balance between exploration and exploitation, 
leading to a superior borrower selection strategy.

Panels (c) and (d) of Figure~\ref{fig:sf_lr10} depict the reward curves and cumulative reward curves for the three agents, respectively. Consistent with the observations from the regret curves, the IDS agent achieves the highest cumulative reward, followed by the THS agent and the GRE agent.  Notably, all agents exhibit positive reward curves throughout the learning process. This implies that even within a borrower population with an expected negative reward, efficient underwriting algorithms can generate profit.

{\flushleft \it Scenario 2: Multiple underwriting with a renewed pool of loan applicants (MR):}\\

We assess agent performance in the MR scenario (refer to Section~\ref{sec:env} for details) which reflects a practical setting with renewed applicant pools and multiple loan approvals per round. The results contrast with the single-context environment (Figure~\ref{fig:singlecontext}).

As observed in the single-context case (Figure~\ref{fig:singlecontext}), the constant flow of new applicants inherently encourages exploration, leading to superior performance by the GRE agent. However, in this scenario with multiple contexts, the non-generalizable nature of the contexts forces GRE to prematurely converge on a suboptimal group of borrowers, resulting in poorer performance.

\begin{figure}[h!]
\begin{center}
\includegraphics[scale=0.35]{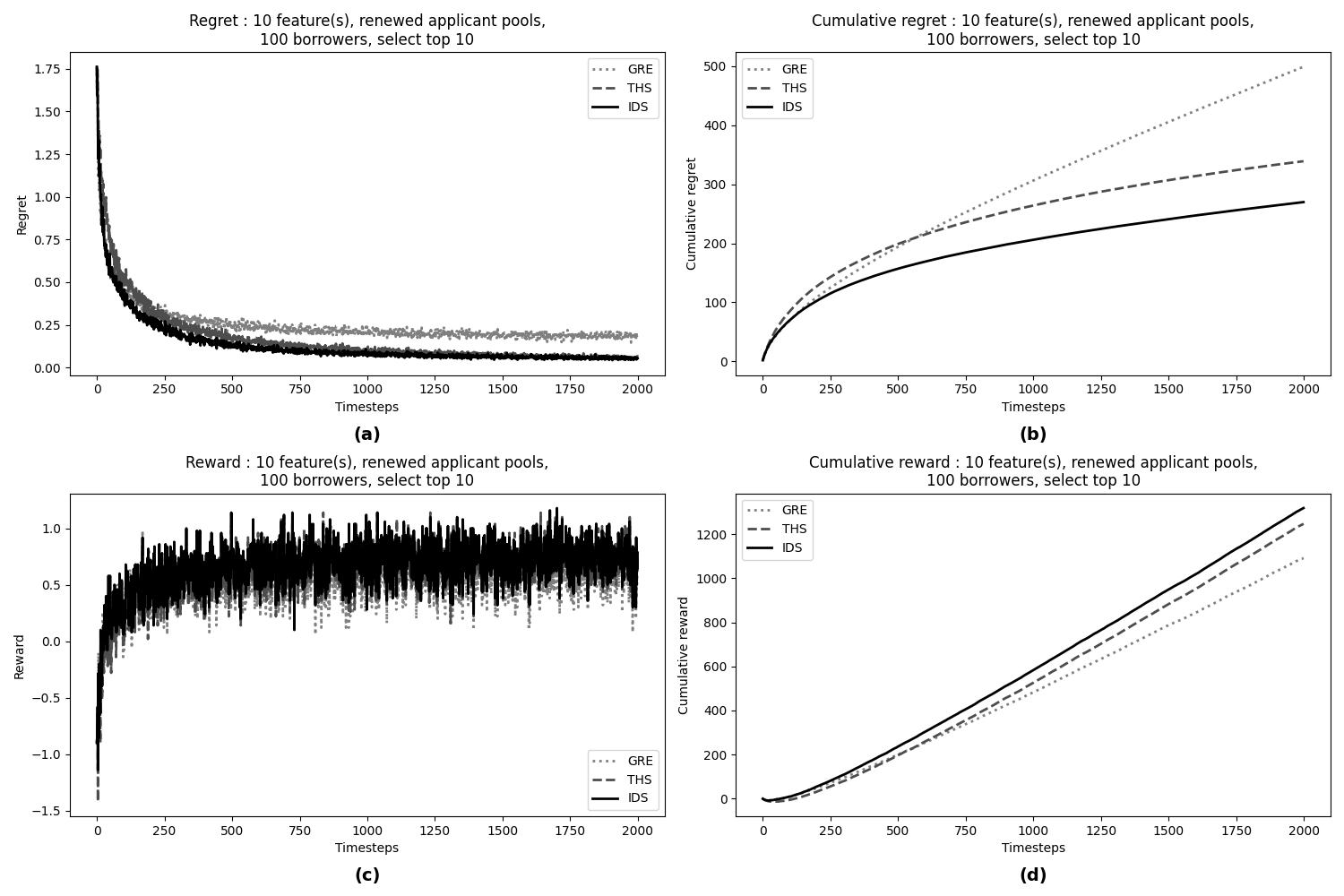}
\caption{The agents' performances with respect to MR scenario (multiple underwriting, renewed applicant pool) 
within the logistic regression environment.  
Panel (a) and (b) exhibit the regret and the cumulative regret curves.  Panel (c) and (d) exhibit the 
reward and cumulative reward curves.}\label{fig:mr_lr10}
\end{center}
\end{figure}

Conversely, agents with more efficient exploration mechanisms, THS and IDS, can more effectively navigate this environment and achieve better results. Figure~\ref{fig:mr_lr10} showcases the performance evaluation of the three agents.

Panels (a) and (b) of Figure~\ref{fig:mr_lr10} depict the regret and cumulative regret, respectively, for each agent. Notably, the regret and cumulative regret for THS and IDS are ultimately lower compared to the GRE agent. This contrasts with the single-context scenario (Figure~\ref{fig:singlecontext}). This finding suggests that the multiple context environment presents a more challenging situation where the inherent exploration driven by new applicant flow is insufficient for GRE's success. In this setting, efficient exploration mechanisms, like those employed by THS and IDS, become crucial for achieving superior performance.

Furthermore, comparing the efficient exploration agents, IDS demonstrates a clear advantage over THS. The faster decrease in regret observed with IDS suggests a superior learning rate towards the true underlying model.

Similar trends are evident in panels (c) and (d) of Figure~\ref{fig:mr_lr10}, which illustrate the expected reward and expected cumulative reward for each agent in the MR scenario. These results further emphasize the superior performance of the agents 
with efficient exploration mechanism. Since all agents exhibit positive reward curves throughout the learning process, the agents exhibit profitability  within a borrower population with an expected negative reward.

\begin{figure}[h!]
\begin{center}
\includegraphics[scale=0.3]{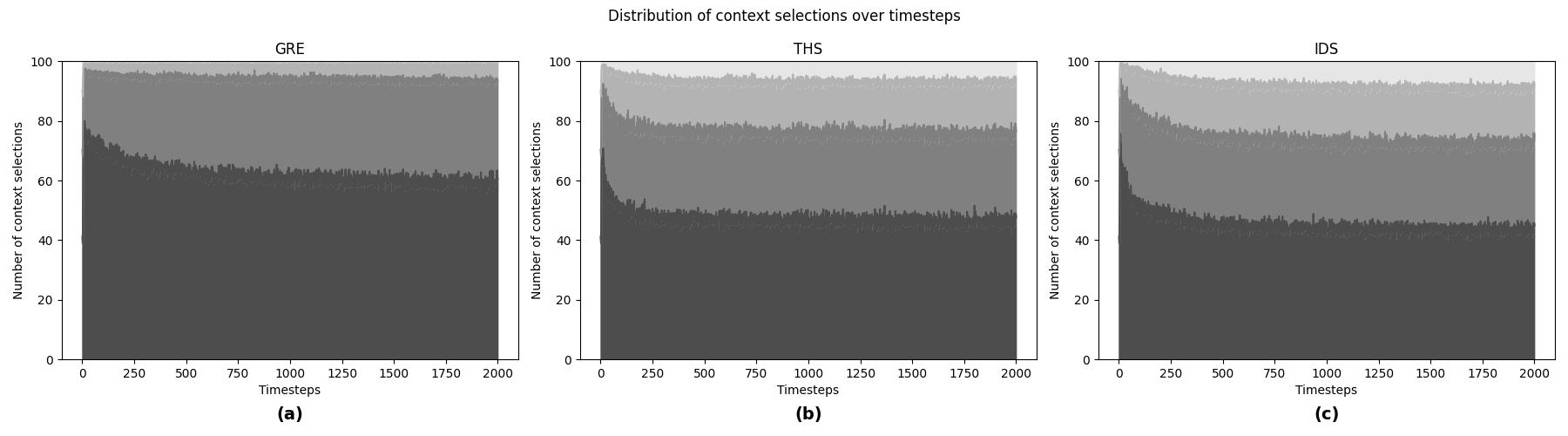}
\caption{Decomposition of loan allocations by borrower group (context) under MR scenario within the logistic regression environment.  
Panel (a): GRE agent allocations.  Panel (b): THS agent allocations. Panel (c): IDS agent allocations.
}\label{fig:alloc}
\end{center}
\end{figure}

Figure~\ref{fig:alloc} visualizes the loan allocation patterns across algorithms. The width of each band in the figure represents the percentage of approved loan applicants originating from a specific borrower context within the approved portfolio of ten loans (averaged over fifty simulations). These bands are ranked by the percentages, 
with the context with the highest percentage at the bottom. 
Notably, the specific borrower context associated with each shade may vary at each time step. Therefore, panels with a greater diversity of distinct shades (e.g., (b) and (c)) suggest a more diversified allocation strategy compared to panels with fewer shades (e.g., (a)). This implies that THS and IDS distribute loans across a wider range of borrower contexts, resulting in more diversified loan portfolios compared to GRE.

\subsection{Environment derived from a neural networks}

This subsection examines an environment where a mismatch exists between the data generation process and the agents' learning model. A neural network model, with ten features ($d=10$) representing borrower characteristics, generates the borrower default data. However, the agents continue to utilize a logistic regression model for credit scoring. This discrepancy between the data's underlying structure and the agents' model stems from the limitations of the latter's capacity. We investigate the performance of the three agents under these challenging conditions.

\begin{figure}[h!]
\begin{center}
\includegraphics[scale=0.75]{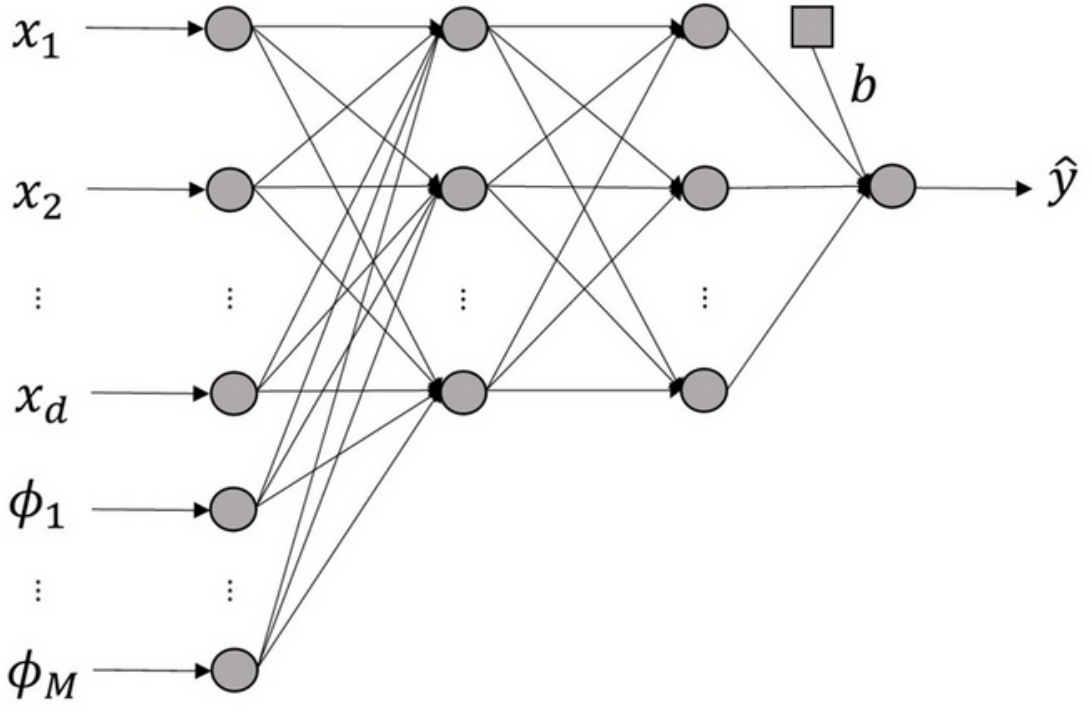}
\caption{Neural network structure for the environment based on a neural network.}\label{fig:nn}
\end{center}
\end{figure}

A fully connected neural network (feedforward) with two hidden layers is chosen as the data generation model. Each hidden layer has the same number of nodes as features ($d$). Nodes in the hidden layer lack bias terms (intercepts), while the output layer's bias 
is added to control the simulated borrowers' average default rate. All nodes, including the output layer, use the sigmoid activation function. Figure~\ref{fig:nn} depicts the chosen neural network structure.

\begin{figure}[h!]
\begin{center}
\includegraphics[scale=0.45]{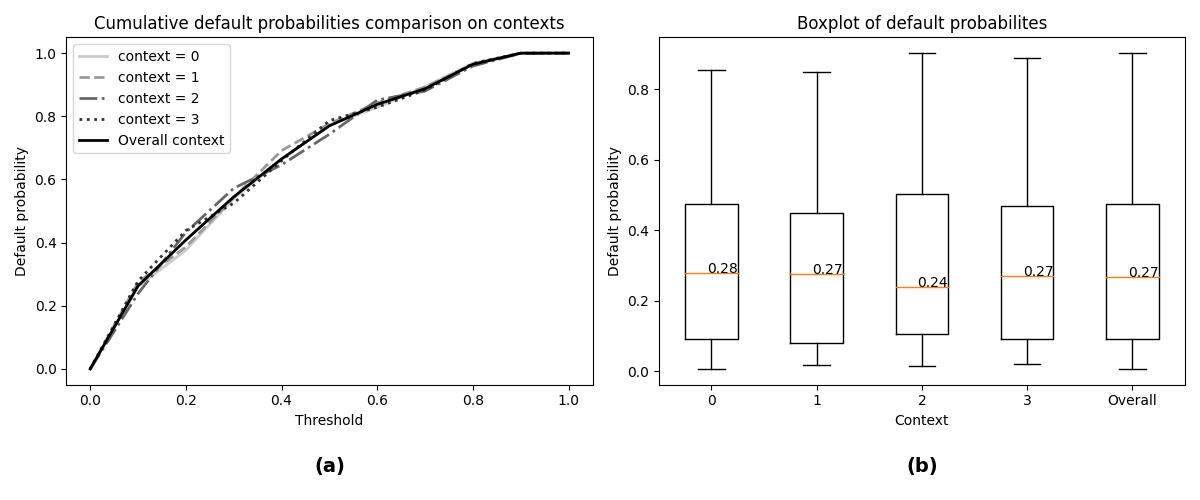}
\caption{The distributions of the default rates by contexts that are simulated under the environment derived from 
the neural network data generation model.  Panel (a) illustrates the distributions by the cumulative distribution function.  
Panel (b) summarizes the distributions in terms of box plots.}\label{fig:dfratenn}
\end{center}
\end{figure}

Borrower feature vectors are drawn from a ten-dimensional multivariate normal distribution with zero mean and identity covariance. Similarly, all weights within the neural network (excluding the output node's bias) are sampled from the same distribution. The output layer's intercept is set to $b=-1.15$, controlling the mean default rate around $27\%$ for each context. Each scenario is simulated fifty times to account for data generation process variability. The resulting distribution of the simulated default rates is visualized as a boxplot in Figure~\ref{fig:dfratenn}. Similar to previous sections, this chosen mean default rate results in a negative population mean reward for the agent with random underwriting algorithm.

Scenarios 3 and 4 correspond to the SF and MR environments described in Section~\ref{sec:env}, respectively. The reported performance metrics represent the average outcome across the fifty simulations

{\flushleft \it Scenario 3: Single underwriting with a fixed pool of loan applicants (SF)}\\

\begin{figure}[h!]
\begin{center}
\includegraphics[scale=0.35]{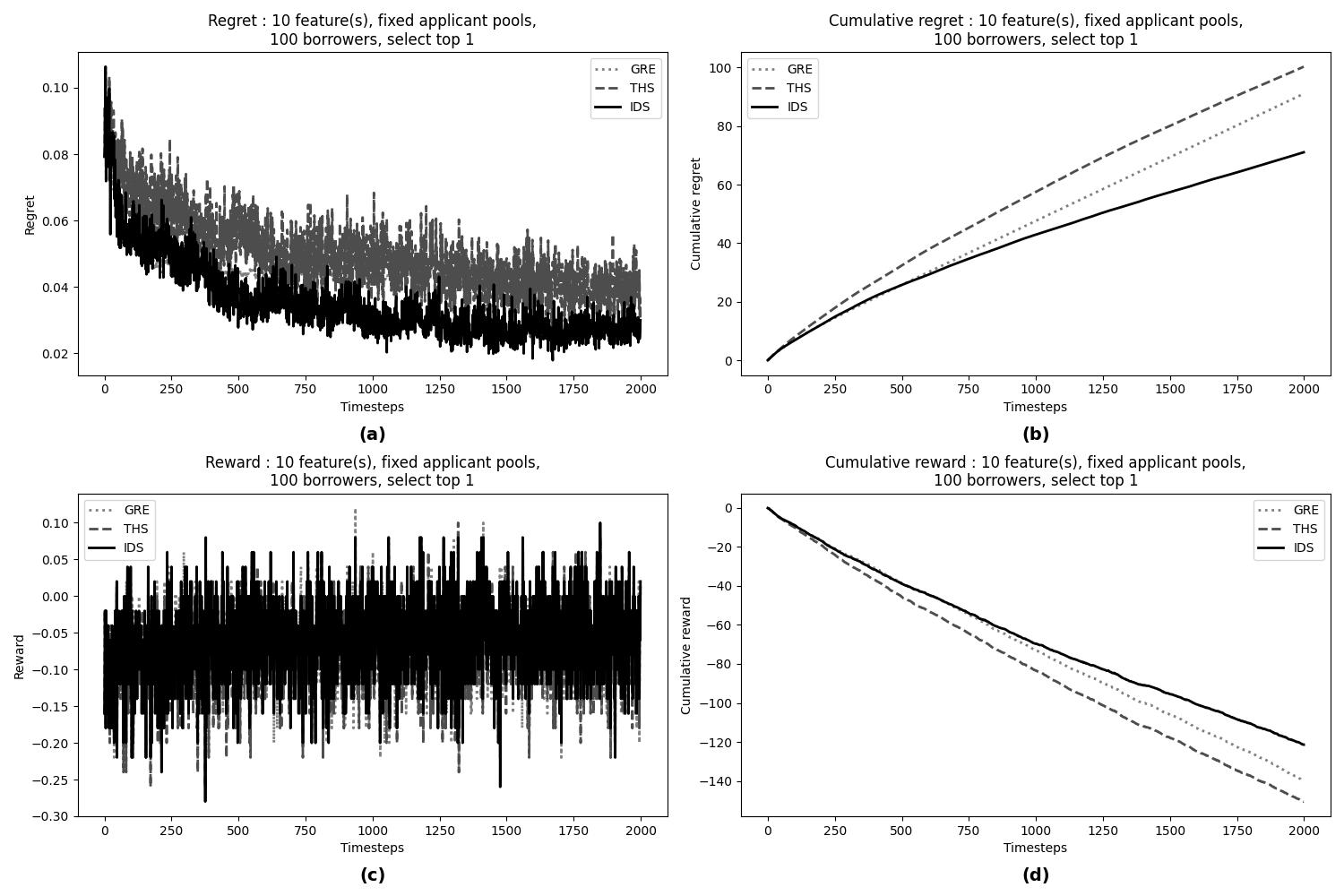}
\caption{The agents' performances with respect to SF scenario (single underwriting, fixed applicant pool) 
within the neural network environment.  
Panel (a) and (b) exhibit the regret and the cumulative regret curves.  Panel (c) and (d) exhibit the 
reward and cumulative reward curves.}\label{fig:sf_nn10}
\end{center}
\end{figure}

In this SF scenario with the neural network-generated data environment, Figure~\ref{fig:sf_nn10} depicts the performance of the three agents: Greedy (GRE), Thompson Sampling (THS), and Information-Directed Sampling (IDS). As before, the tuning parameter $p$ for the IDS agent remains set at $3$.

Panel (a) of Figure~\ref{fig:sf_nn10} reveals that all regret curves remain well above zero, indicating the agents' limited ability to fully learn the environment. This is understandable given the discrepancy between the true environment's data generation model and the agents' learning models.

In contrast to prior findings, all performance metrics consistently show that THS performs the worst among the three agents. Between the remaining two, IDS maintains its lead over GRE. These results suggest that when agents have limited learning capacity, an efficient exploration-exploitation strategy might not always outperform a pure exploitation strategy.

Panels (c) and (d) highlight a crucial point: even with efficient exploration (THS and IDS), none of the agents could achieve profitability within this challenging environment. This is because, as previously mentioned, the expected reward under the default rate distribution is negative, making profitability inherently difficult.  This reinforces the notion that, in competitive environments, learning the correct underlying model is essential for achieving profitability.

{\flushleft \it Scenario 4: Multiple underwriting with a renewed pool of loan applicants (MR):}\\

\begin{figure}[h!]
\begin{center}
\includegraphics[scale=0.35]{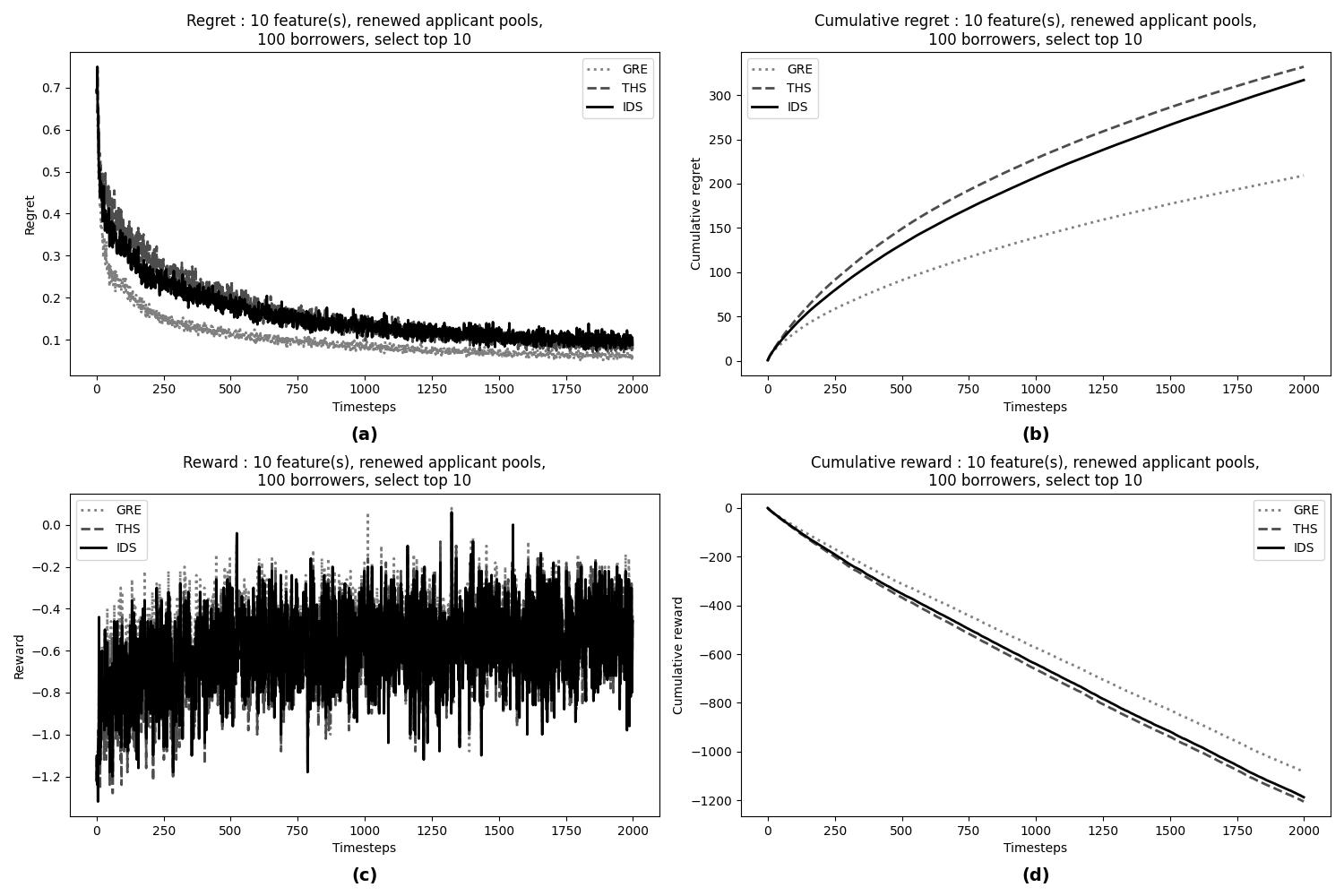}
\caption{The agents' performances with respect to MR scenario (multiple underwriting, renewed applicant pool) 
within the neural network environment.  
Panel (a) and (b) exhibit the regret and the cumulative regret curves.  Panel (c) and (d) exhibit the 
reward and cumulative reward curves.}\label{fig:mr_nn10}
\end{center}
\end{figure}

In the MR scenario, employing the same neural network-generated data environment, Figure~\ref{fig:mr_nn10} shows the performance of the three agents. We maintain the tuning parameter $p$ at 3 for consistent comparison across scenarios.

Consistent with Scenario 3, Panels (a) and (b) demonstrate that 
all agents struggle to learn the environment effectively due to the model's limited capacity. However, unlike previous scenarios, the GRE agent outperforms THS and IDS agents in this MR setting. This suggests that under learning model's limited capacity, the inherent exploration driven by the constant influx of new applicants outweighs the benefits of the additional exploration strategies employed by THS and IDS.

As shown in Panels (c) and (d), none of the agents achieve profitability. This further reinforces the notion that learning the correct underlying model is critical for strong performance.

\begin{figure}[h!]
\begin{center}
\includegraphics[scale=0.3]{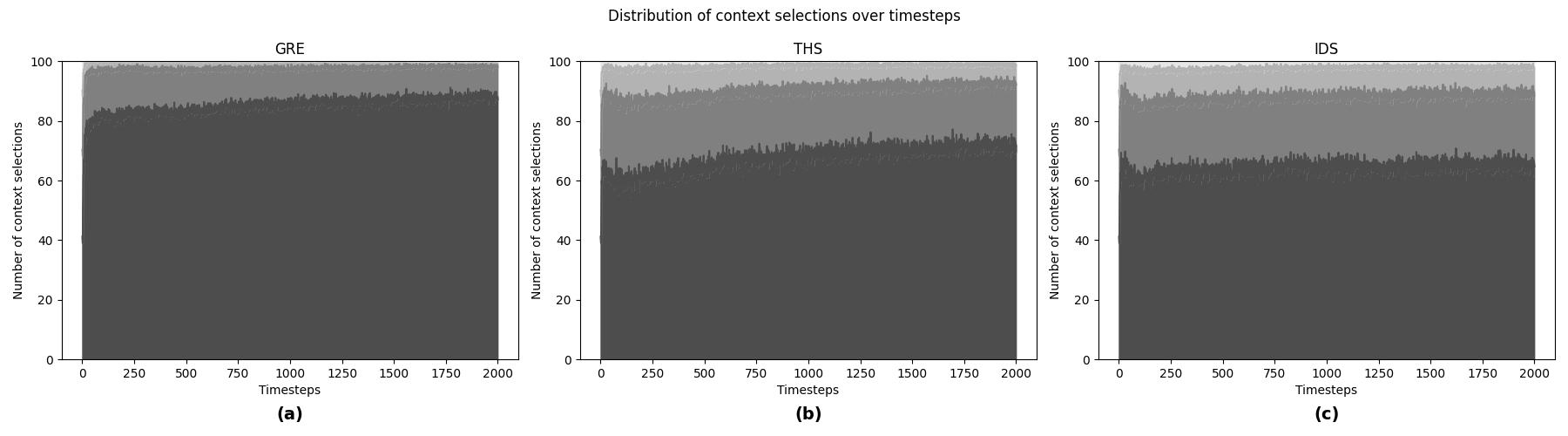}
\caption{Decomposition of loan allocations by borrower group (context) under MR scenario within the neural network environment.  
Panel (a): GRE agent allocations.  Panel (b): THS agent allocations. Panel (c): IDS agent allocations.}\label{fig:alloc_nn}
\end{center}
\end{figure}

Figure~\ref{fig:alloc_nn} illustrates the loan allocation patterns across algorithms. Notably, the loan portfolios of each agent in this scenario exhibit less diversification compared to Scenario 2. This implies that an overly simplistic learning model not only hinders profitability but also restricts the agents from making diverse choices.

\section{Conclusion}

This paper presents a novel credit underwriting framework that integrates credit scoring seamlessly into the loan approval process, specifically targeting the challenge of ungeneralizable contextual credit scoring. This framework leverages the principles of reinforcement learning (RL) with key modifications, including action space renewal and multi-choice actions, while restricting ourselves to a logistic regression learning model. We demonstrate that by framing the underwriting process as an RL problem, the traditional underwriting approach aligns with the RL greedy algorithm. We further propose two novel and more sophisticated algorithms based on Thompson Sampling and Information Directed Sampling.

Our findings reveal that the new algorithms offer significant performance improvements over the traditional approach in ungeneralizable contextual credit scoring scenarios where the environment's data aligns with the agent's learning model 
(i.e., a logistic regression model). Additionally, these sophisticated algorithms exhibit a more diversified decision-making pattern compared to the traditional method. However, in scenarios where the underlying data is significantly more complex than the agent's model, the exploration benefits introduced by action space renewal enable the traditional algorithm to outperform the more sophisticated ones.

This observation highlights a crucial point: while advanced exploration strategies are valuable, they cannot fully compensate for an agent's inability to learn an accurate model of the environment. This underscores the importance of employing powerful machine learning models and methodologies within credit underwriting frameworks, especially when dealing with complex data structures. A limitation of the current approach is its reliance on a single logistic regression model. Neural network models, with their ability to capture complex data structures, present a promising avenue for improving model accuracy, especially when coupled with efficient exploration mechanisms~\citep{osband2023epistemic}. Therefore, the development of neural network models specifically designed for efficient exploration in ungeneralizable contextual credit scoring scenarios holds immense promise for significantly advancing the accuracy and effectiveness of RL-based credit underwriting frameworks. Further investigation into this area is warranted.


\bibliographystyle{plainnat}
\bibliography{creditrl}

\end{document}